\documentclass[letterpaper]{article} 
\usepackage{aaai25}  
\usepackage{times}  
\usepackage{helvet}  
\usepackage{courier}  
\usepackage[hyphens]{url}  
\usepackage{graphicx} 
\usepackage{natbib}  
\usepackage{caption} 
\usepackage{algorithm}
\usepackage{algorithmic}

\setcitestyle{authoryear,open={(},close={)},citesep={,},maxcitenames=1,mincitenames=1}

\usepackage{enumitem}
\usepackage{multirow}
\usepackage{multicol}
\usepackage{rotating}

\usepackage{makecell}
\usepackage{colortbl}
\usepackage{xcolor}
\definecolor{arrowred}{RGB}{255,0,0}
\usepackage{amssymb}

\usepackage{dsfont}
\usepackage{url}
\usepackage{tabularx,booktabs}
\usepackage{bbding}
\usepackage{pifont}
\usepackage[nointegrals]{wasysym}
\usepackage{adjustbox}

\usepackage{amsmath}

\usepackage{cleveref}
\crefname{figure}{Fig.}{Figures}
\crefname{table}{Tab.}{Tables}
\crefname{equation}{Eq.}{Equation}

\usepackage{caption}

\usepackage{newfloat}
\usepackage{listings}
\DeclareCaptionStyle{ruled}{labelfont=normalfont,labelsep=colon,strut=off} 
\floatstyle{ruled}
\newfloat{listing}{tb}{lst}{}
\floatname{listing}{Listing}
\title{\textbf{LiON}: \underline{L}earning Po\underline{i}nt-wise  Abstaining Penalty for LiDAR  \underline{O}utlier Detectio\underline{N} Using Diverse Synthetic Data}
\author{
    Shaocong Xu\textsuperscript{\rm 1,2}\equalcontrib, Pengfei Li\textsuperscript{\rm 1}\equalcontrib, Qianpu Sun\textsuperscript{\rm 1}, Xinyu Liu\textsuperscript{\rm 1}, Yang Li\textsuperscript{\rm 1}, Shihui Guo\textsuperscript{\rm 2}\thanks{Corresponding author.}, Zhen Wang\textsuperscript{\rm 3}, \\Bo Jiang\textsuperscript{\rm 3}, Rui Wang\textsuperscript{\rm 3}, Kehua Sheng\textsuperscript{\rm 3}, Bo Zhang\textsuperscript{\rm 3}, Li Jiang\textsuperscript{\rm 4}, Hao Zhao\textsuperscript{\rm 1}\footnotemark[2], Yilun Chen\textsuperscript{\rm 1}
}
\affiliations{
    \textsuperscript{\rm 1}Tsinghua University, \textsuperscript{\rm 2}Xiamen University, \textsuperscript{\rm 3}Didi Chuxing, \textsuperscript{\rm 4}Chinese University of Hong Kong, ShenZhen
}

\usepackage{bibentry}

\begin{document}

\maketitle

\begin{abstract}
LiDAR-based semantic scene understanding is an important module in the modern autonomous driving perception stack. However, identifying outlier points in a LiDAR point cloud is challenging as LiDAR point clouds lack semantically-rich information. While former SOTA methods adopt heuristic architectures, we revisit this problem from the perspective of Selective Classification, which introduces a selective function into the standard closed-set classification setup. Our solution is built upon the basic idea of abstaining from choosing any inlier categories but learns a point-wise abstaining penalty with a margin-based loss. Apart from learning paradigms, synthesizing outliers to approximate unlimited real outliers is also critical, so we propose a strong synthesis pipeline that generates outliers originated from various factors: object categories, sampling patterns and sizes. We demonstrate that learning different abstaining penalties, apart from point-wise penalty, for different types of (synthesized) outliers can further improve the performance. We benchmark our method on SemanticKITTI and nuScenes and achieve SOTA results. Codes are available at \url{https://github.com/Daniellli/LiON/}.
\end{abstract}


%

\section{Introduction}


LiDAR outlier detection \cite{li2023open} complements LiDAR semantic segmentation \cite{wang2024sfpnet}, aiming to enhance the model’s ability to recognize outliers without compromising its inlier segmentation performance.

\begin{figure}[t]
\centerline{\includegraphics[width=0.45\textwidth]{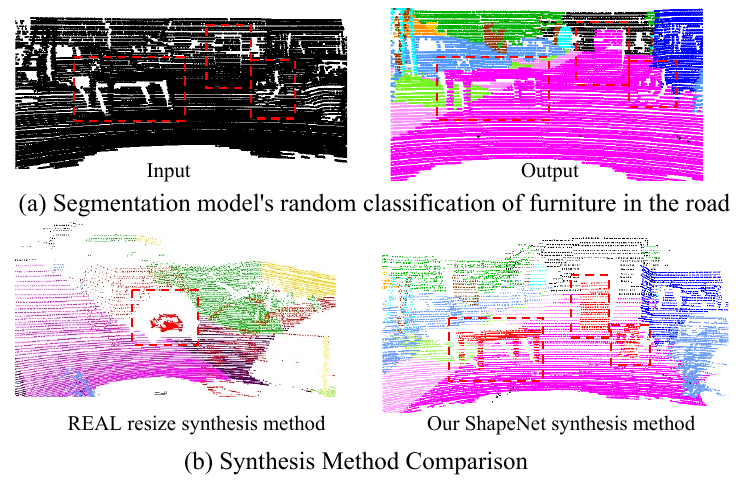}}
\vspace{-1.em}
\caption{(a). The semantic segmentation model fails to identify furniture because the training set does not include such objects; (b). Comparison of our ShapeNet outlier synthesis method and the former resize outlier synthesis method.}
\label{fig:teaser}
\vspace{-1.5em}
\end{figure}

This task is important and practical. Traditional segmentation methods \cite{wang2024sfpnet,li2024rapid} assume that the samples in the training and test sets belong to the same set of categories. Thus, these models are trained on inlier categories and tend to classify all inputs into one of the inlier categories. However, this assumption fails in real-world scenarios where outliers are present. For example, as shown in \cref{fig:teaser}-(a), the model may randomly classify furniture that has not been seen in training, leading to disastrous consequences in the downstream planning stage.

While 2D outlier detection has made significant strides, including optimizing inlier prediction \cite{tian2022pixel, liu2023residual, miao2024out}, addressing outlier class imbalance \cite{choi2023balanced}, evolving from pixel-wise to mask-based outlier detection methods \cite{zhang2024csl, rai2023unmasking, nayal2023rba, zhang2024csl}, developing promptable outlier detectors \cite{zhao2024segment, li2024learning, zhou2022learning}, and utilizing model ensembling \cite{liu2024ensemble}, the field of LiDAR outlier detection is still in its early stages \cite{li2023open}. Seminal work REAL \cite{cen2022open} proposes randomly choosing and resizing objects existing in the scene to synthesize outliers to approximate unlimited real outliers. For example, as shown in the left panel of \cref{fig:teaser}-(b), the car is chosen and shrunk. This is viable but fail to represent the long-tail distribution of real outliers, in two regards. Firstly, objects from existing road scene understanding datasets are limited in category. Secondly, naive resizing violates the sampling pattern of real LiDAR sensors: points on enlarged objects get sparser while those on shrunk objects get denser. This leads to a shortcut problem: the model may find a trivial solution to distinguish outliers from inliers solely using point sparsity.

Besides, REAL empirically finds 2D outlier detection methods \cite{hendrycksBaselineDetectingMisclassified2018, hendrycks2019scaling, gal2016dropout} perform poorly in the 3D domain, as a large number of outliers are predicted as inliers with high confidence scores . To alleviate this phenomenon, in addition to the Cross Entropy (CE) loss, REAL designs a Calibration Cross Entropy(CCE) loss to calibrate the outlier probabilities in inlier prediction. However, we empirically find that CCE alleviates this at the cost of accuracy in predicting inliers. As shown in the right panel of \cref{fig:real_statistics}, with CCE, the outlier probabilities of inlier points, which account for the vast majority of the total points, become very high (higher than 0.1), which is undesirable.

To address these issues, in this work, we propose a novel method LiON, aiming to mitigate the lack of semantically-rich information in LiDAR point clouds for outlier detection. We contribute from two perspectives: learning and data.


\textbf{Learning.} We reformulate the LiDAR outlier detection problem by applying Selective Classification (SC) principles \citep{feng2023towards, chow1970optimum} and introduce a point-wise abstaining penalty learning paradigm to address the problem of unclear distinction in point clouds. While inspired by SC, our method differs significantly from SC in our point-wise design. Specifically, a diverse calibration factor is learned in a point-wise manner to more effectively capture subtle differences within a point cloud and calibrate the relationship between inlier and outlier classifiers. As a result, we mitigate the unclear distinction in point clouds caused by their lack of semantically-rich information by learning a diverse factor in a point-wise manner.

\textbf{Data.} Inspired by outlier exposure \cite{hendrycks2019oe}, we propose introducing objects from an external dataset, ShapeNet \citep{chang2015shapenet}, into existing scenes to synthesize outliers. ShapeNet, with its wide spectrum of categories and diverse geometries, can compensate for the long-tail distribution of real outliers. To ensure the realism of synthesized outliers, we take the LiDAR sampling pattern into consideration when merging randomly selected ShapeNet objects into road scenes. As illustrated in the right of \cref{fig:teaser}-(b), our synthesized outliers are precisely aligned with objects in the scene with respect to point sparsity and occlusion. In this way, we mitigate the lack of semantically-rich information in LiDAR point clouds by utilizing realistic, ShapeNet outliers with diverse geometries.

Finally, the risk-coverage evaluation metrics associated with SC are also adapted for this task to serve as supplementary metrics of the holistic metrics AUPR/AUROC/mIoU$_\text{old}$ , allowing us to gain a deeper understanding of the performance gains. These metrics are also a key to narrow the gap between academic and industrial communities. This is because these metrics allow us to identify the rejection threshold that incurs the least cost but yields the highest gain (coverage), which is very important for industrial applications. 


Our contributions can be summarized as follows:
\begin{itemize}
    \item We propose a point-wise abstaining penalty learning paradigm using the principle of SC to calibrate the relationship between inlier and outlier classifiers in a point-wise manner. Additionally, the risk-coverage evaluation metrics associated with SC are adapted for this problem, serving as supplementary metrics to the holistic metrics AUPR/AUROC/mIoU$_\text{old}$.
    
    \item We utilize ShapeNet objects to synthesize realistic and diverse outliers to approximate unlimited real outliers.
    
    \item Our method has achieved new SOTA performance for LiDAR outlier detection not only in previously established outlier detection metrics, but also in the risk-coverage curve metric, on SemanticKITTI and NuScenes. 
    
\end{itemize}

\section{Related Works}

\textbf{Outlier Detection in Autonomous Driving.} Outlier detection is vital for ensuring the safety of ego-cars in open-world environments by identifying outlier objects. Extensive research has been conducted in 2D perception, specifically with semantic segmentation models. Unsupervised methods \citep{jung2021standardized,bevandic2021dense} involve post-processing predicted logits from frozen segmentation models to detect outliers. Supervised methods \citep{tian2022pixel,grcic2022densehybrid,chan2021entropy} utilize auxiliary datasets like COCO \citep{lin2014microsoft} to synthesize outlier objects in training images (e.g., Cityscapes \citep{cordts2016cityscapes}) and retrain the segmentation model using outlier exposure \citep{hendrycks2019oe, zhou2024pad}.

While significant advancements have been made in 2D outlier detection, the exploration in the context of LiDAR point clouds remains limited. \citet{cen2022open} use resize synthesis pipeline and calibration loss to achieve the discrimination of outlier points. However, their focus primarily revolves around the open-world segmentation setting without extensively analyzing calibration effectiveness. \citet{li2023open} propose an adversarial prototype framework that improves performance but involves complex network design and computationally expensive training. Considering these limitations, we present our method and substantiate its effectiveness through extensive experiments.

\textbf{Selective Classification.} SC can be broadly classified into two groups: 1) the first group focuses on addressing SC through the use of additional heads/logits \citep{geifman2017selective,geifman2019selectivenet,liu2019deep,feng2023towards,gal2016dropout,chow1970optimum}; 2) the second group tackles SC through cost-sensitive classification techniques \citep{charoenphakdee2021classification,mozannarConsistentEstimatorsLearning2020}. Motivated by the extra head/logits design, resembling outlier detection, we design a new outlier detector from the perspective of SC and utilize SC's evaluation metrics as alternative performance measures for our outlier detector.

\textbf{Synthetic Data.} Due to the limited availability and high cost associated with acquiring high-quality real-world datasets, synthetic data has been extensively employed in the field of scene understanding \citep{suRenderCNNViewpoint2015, movshovitz-attiasHowUsefulPhotorealistic2016,zhangPhysicallyBasedRenderingIndoor2017,handa2016understanding,mccormac2017scenenet,zhang2017deepcontext, song2017semantic, gao2021rcd,gao2024zero,gao2024collaborative,gao2023leveraging,xu2024ect, ding2024preafford}. In this work, we introduce ShapeNet objects into the existing scene to effectively compensate for the lack of semantically-rich information in point clouds.
\begin{figure}[t]
\centerline{\includegraphics[width=0.5\textwidth]{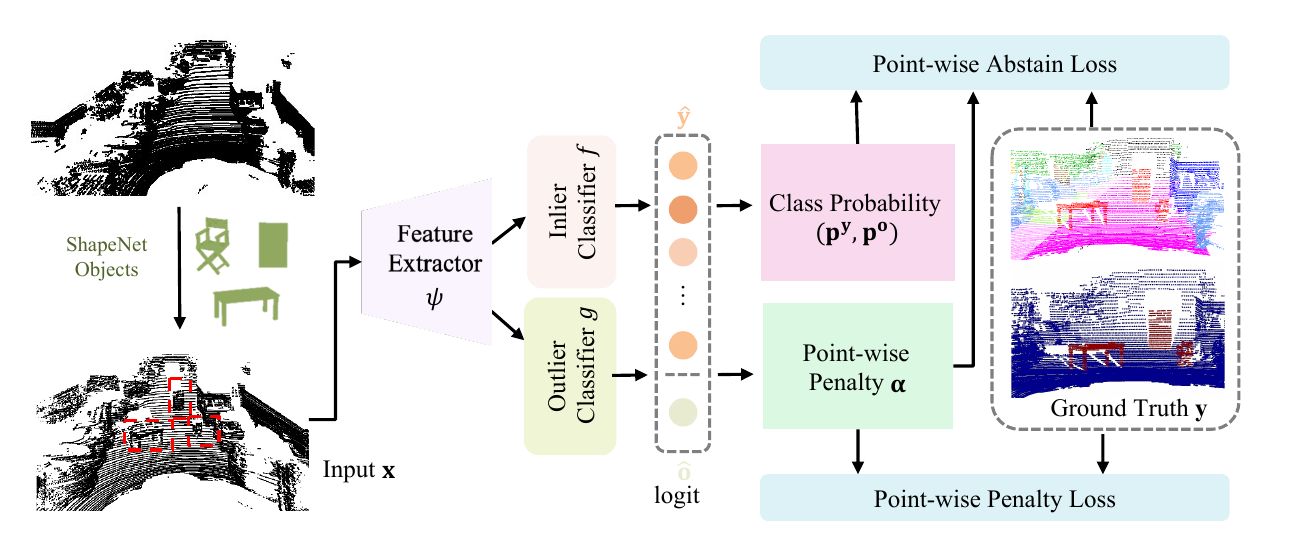}}
\caption{Method pipeline: a point cloud containing outliers synthesized by ShapeNet is processed by a feature extractor to obtain features. These features are then used by inlier and outlier classifiers to predict class logits.}

\label{fig:main}
\vspace{-1em}
\end{figure}

\section{Preliminaries: Selective Classification}
\label{section:sc_definition}

We first formalize the definition of SC and put different methods under a unified lens. 
Experimentally, the risk-coverage trade-off analysis that comes along with this framework, allows us to reveal in-depth differences between methods.

\textbf{Definition.} In SC, our goal is to learn predictive models that know what they do not know or when they should abstain\footnote{The abstaining penalty will be defined later.} from making decisions. Here we consider a generic SC definition, which is agnostic of network and application. A supervised classification task is formulated as follows. Let $\mathcal{X}$ be any feature space and $\mathcal{Y}$ a label space. In LiDAR outlier detection, $\mathcal{X}$ could be point clouds, and $\mathcal{Y}$ could be class labels\footnote{These class labels include an outlier label.} of each point cloud. Let $P(\mathcal{X},\mathcal{Y})$ be a distribution over $\mathcal{X} \times \mathcal{Y} $. A model $f: \mathcal{X} \rightarrow \mathcal{Y}$ is called a prediction function and its true risk used to evaluate the performance of $f$ w.r.t. $P$ is $R(f) := E_{P(\mathcal{X},\mathcal{Y})}[\ell(f(\mathbf{x}),\mathbf{y})] $, where $\ell: \mathcal{Y} \times \mathcal{Y} \rightarrow \mathbb{R}^+$ is a given loss function, for example, the Cross Entropy (CE) loss. Given a labeled set $S_m = \{(\mathbf{x_i}, \mathbf{y_i})\}_{i=1}^m$ sampled i.i.d. from $P(\mathcal{X}, \mathcal{Y})$, the empirical risk of the classifier $f$ is $\hat r(f | S_m) := \frac{1}{m} \sum_{i=1}^{m}\ell (f(\mathbf{x_i}),\mathbf{y_i})$.

\textbf{Apart from risk}, another important concept in the SC formulation is coverage. A selective model \citep{el2010foundations} is a pair $(f,g)$, where $f$ is a prediction function, and $g: \mathcal{X} \rightarrow \{0,1\}$ is a selective function, which serves as a binary qualifier for $f$ as follows:
\begin{equation}
\begin{aligned}
    (f,g)(\mathbf{x}) := 
    \begin{cases}
        f(\mathbf{x}), & \text{if } g(\mathbf{x}) = 1 \\
        \text{ABSTAIN}, & \text{if } g(\mathbf{x}) = 0 \\
    \end{cases}
  \end{aligned}
  \label{eq:selctive formulation}
\end{equation}
Thus, the selective model abstains from prediction at $\mathbf{x} \text{ iff } g(\mathbf{x}) = 0$. A soft selection function can also be considered, where $g: \mathcal{X} \rightarrow [0, 1]$, and decisions can be taken probabilistically or deterministically (e.g., using a threshold). The introduction of a selective function $g$ allows us to define coverage. Specifically, coverage is defined to be the ratio of the non-abstained subset within set $P$ to the entirety of $P$, which can be formulated as: 
\begin{equation}
\begin{aligned}
    \phi(g) := E_P[g(\mathbf{x})]
  \end{aligned}
  \label{eq:coverage formulaiton}
\end{equation}
Accordingly, the standard risk for a classifier $f$ can be augmented into the selective risk of $(f, g)$ as
\begin{equation}
\begin{aligned}
    R(f,g) := \frac{E_P[{\ell(f(\mathbf{x}),\mathbf{y})g(\mathbf{x})}]}{\phi(g)}
  \end{aligned}
  \label{eq:risk formulation1}
\end{equation}
Clearly, the risk of a selective model can be traded-off for coverage. The performance profile of such a model can be specified by its risk–coverage curve, defined to be the risk as a function of coverage.

Finally, we clarify that the continuous risk and coverage defined above are calculated using a fixed set in practice. For any given labeled set $S_m$, the empirical selective risk is
\begin{equation}
\begin{aligned}
    \hat r(f,g|S_m) := \frac{ \frac{1}{m}\sum_{i=1}^m \ell(f(\mathbf{x_i}), \mathbf{y_i})g(\mathbf{x_i})}{\phi(g|S_m)}
  \end{aligned}
  \label{eq:empirical-risk-formulation-definition}
\end{equation}
and the empirical coverage is
\begin{equation}
\begin{aligned}
    \hat \phi(g|S_m) := \frac{1}{m} \sum_i^m g(\mathbf{x_i})
  \end{aligned}
  \label{eq:empirical coverage}
\end{equation}
\textbf{Remark.} Later we will put methods under the unified lens of SC and use the risk-coverage curve to analyze them.
\section{Method}
\label{section:method}
LiDAR semantic segmentation is the task of assigning a class label from a predefined class label set to each point in a given point cloud. LiDAR outlier detection, on the other hand, is an extension to semantic segmentation, which aims to identify points that do not belong to the predefined inlier label set. While the seminal work proposes a viable solution REAL, we revisit this issue through the unified lens of SC and propose a solution that effectively mitigates the problem of unclear distinctions in point clouds. Apart from the learning paradigm, we design a novel outlier synthesis pipeline that leverages the richness of the ShapeNet and adheres to the realistic LiDAR distribution to compensate for the lack of semantically-rich information in point clouds.

\subsection{Network Architecture Overview}
As shown in \cref{fig:main}, the input point cloud $\mathbf{x} \in \mathbb{R}^{n\times3}$, sampled from $S_m$, is denoted on the left with ShapeNet objects integrated into it. Then, $\mathbf{x}$ is fed into the feature extractor $\psi$ followed by an inlier classifier $f$ to predict the inlier logit $\mathbf{\hat{y}} \in \mathbb{R}^{n \times c}$, where $c$ is the number of inlier classes. An outlier classifier $g$ is used to predict the outlier logit $\mathbf{\hat{o}} \in \mathbb{R}^{n \times 1}$. As such, $(f, g)$ instantiates a selective model mentioned above. These operations can be expressed as follows:
\begin{equation}
\begin{gathered}
    \mathbf{\hat{y}}  := f(\psi(\mathbf{x})) \quad
    \mathbf{\hat{o}} := g(\psi(\mathbf{x})) \\
    \mathbf{\tilde{y}}  := [\mathbf{\hat{y}}, \mathbf{\hat{o}}] := \Biggl\{\mathbf{\tilde{y}_{i}} := [\hat{y}_i,\hat{o}_i] \Big| i=1,\ldots,n\Biggr\} \\
     \mathbf{p} :=\Biggl \{p_{i,j} = \frac{e^{\tilde{y}_{i,j}}}{\sum_{k = 1}^{c+1} e^{\tilde{y}_{i,k}}} \Big| i = 1, \ldots, n; j = 1, \ldots, c + 1 \Biggr\} \\
     \mathbf{p^y}, \mathbf{p^o} := \mathbf{p}
  \end{gathered}
  \label{eq:pipeline formulation}
\end{equation}
Here, the operation $[\cdot]$ denotes concatenation. Prediction probability $\mathbf{p}\in [0,1]^{n\times(c+1)}$ consists of inlier probability $\mathbf{p^y} \in [0,1]^{n \times c}$ and outlier probability $\mathbf{p^o} \in [0,1]^{n \times 1}$.


\subsection{Revisiting the REAL Formulation}

\citet{cen2022open} introduces the first LiDAR outlier detector, REAL. Their added dummy classifiers can also be thought of as $g$, under the SC framework, but there is a key difference. They observe that numerous real outliers are wrongly classified as inlier classes with high probabilities. To address this issue, they propose a Calibration Cross Entropy (CCE) loss function to drive the outlier logit of the inlier sample to the second largest. We formalize\footnote{This is equivalent to the loss used in the REAL paper although REAL describes it in a different manner.} this loss as follows:
\begin{equation}
\begin{gathered}
\ell := \frac{1}{m}\sum_{S_m} \frac{1}{n}\sum_{i=1}^n \Biggl\{ \underbrace{- \log{
p_{i,y_i}
}
}_\text{CE} \\
\underbrace{- \lambda \mathbb{I}(y_i \neq c+1) \log{\frac{e^{\tilde{y}_{i,c+1}} }{\sum_{k=1 \& k \neq y_i}^{c+1}e^{\tilde{y}_{i,k}}}}}_\text{CCE} \Biggr\}\\
  \end{gathered}
  \label{eq:REAL-formulation}
\end{equation}
Here, $y_i \in \{1,\ldots,c,c+1\}$ signifies the ground truth corresponding to $x_i \in \mathbf{x}$, where $\{1,\ldots,c\}$ represents inlier class labels while $\{c+1\}$ indicates the outlier class label. $\mathbf{y}$ and $\mathbf{x}$ are sampled i.i.d. from $S_m$. $\mathbb{I}(\cdot)$ is the indicator function. $\lambda$ is a hyperparameter. A notable fact is that REAL does not provide an ablation study for this CCE loss and as shown in ~\cref{tab:REAL ablation }, removing this CCE loss can indeed improve standard outlier detection metrics. But why this happens cannot be understood through metrics like AUPR or AUROC, highlighting the need to revisit REAL under the lens of SC. 

\begin{figure}[t]
\centerline{\includegraphics[width=0.45\textwidth]{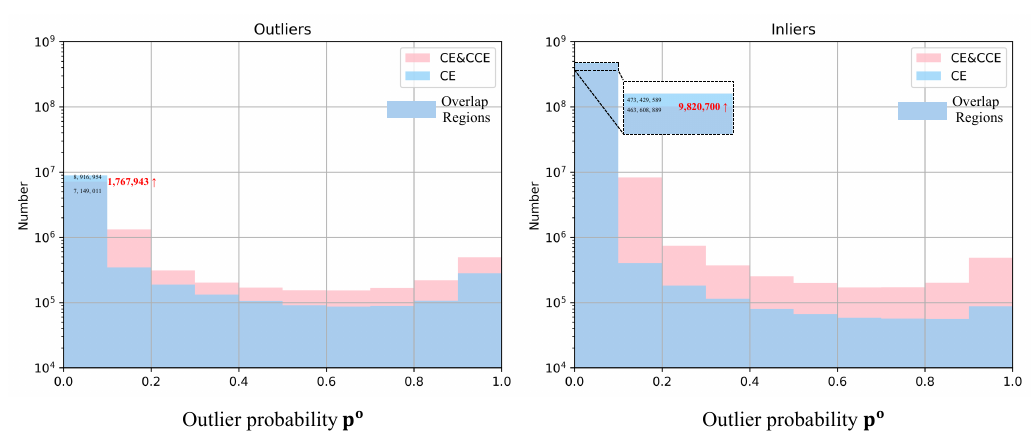}}
\vspace{-1em}
\caption{Statistics of outlier probability $\mathbf{p^o}$ for inlier and outlier samples under different settings of REAL on SemanticKITTI. For inliers, we would like to observe more samples with $\mathbf{p^o} \in [0,0.1]$, while for outliers, less samples with $\mathbf{p^o} \in [0,0.1]$ is desirable.}
\label{fig:real_statistics}
\vspace{-1.5em}
\end{figure}

We use \cref{fig:real_statistics} as an intuitive case to analyze the negative impact of the CCE loss. It illustrates sample numbers within different $\mathbf{p^o}$ intervals for both inliers and outliers. By removing the CCE loss, sample number in the extremely low interval $[0,0.1]$ grows significantly for both inliers and outliers. This is desirable for inliers but not for outliers. As shown by the red increase number, 9.8 million samples change to a desirable state while 1.7 million samples change to a undesirable state, and this large difference explains why the collective metrics in \cref{tab:REAL ablation } become better.

The reason why this in-depth statistics (\cref{fig:real_statistics}) can reveal the negative impact of CCE is that different $\mathbf{p^o}$ values are investigated separately. And, the SC framework provides the principled tool risk-coverage curve to conduct this kind of analysis, because different coverage is achieved through selecting different thresholds on $\mathbf{p^o}$. More principled analyses that reveal reasons behind phenomena like \cref{tab:REAL ablation } can be found in the experiments section.

\begin{table}[t]
  \centering
    \vspace{-0.5em}
  \centering
  \begin{adjustbox}{width=0.3\textwidth}
    \begin{tabular}{c|c|c|c|c}
    \toprule
      $\text{CE} $ & CCE & AUPR & AUROC & $\text{mIoU}_{\text{old}}$  \\
     \midrule
     \midrule
      \checkmark & \checkmark & 20.00 & 84.90 & 57.80 \\
    \checkmark & & \textbf{26.68} & \textbf{87.60} & \textbf{58.28} \\
    \bottomrule
    \end{tabular}%
    \end{adjustbox}
  \caption{ Removing CCE in REAL can actually improve the outlier detection metrics on SemanticKITTI.}
  \label{tab:REAL ablation }
\vspace{-.5em}
\end{table}

\begin{figure*}[t]
\centering
\includegraphics[width=0.8\textwidth]{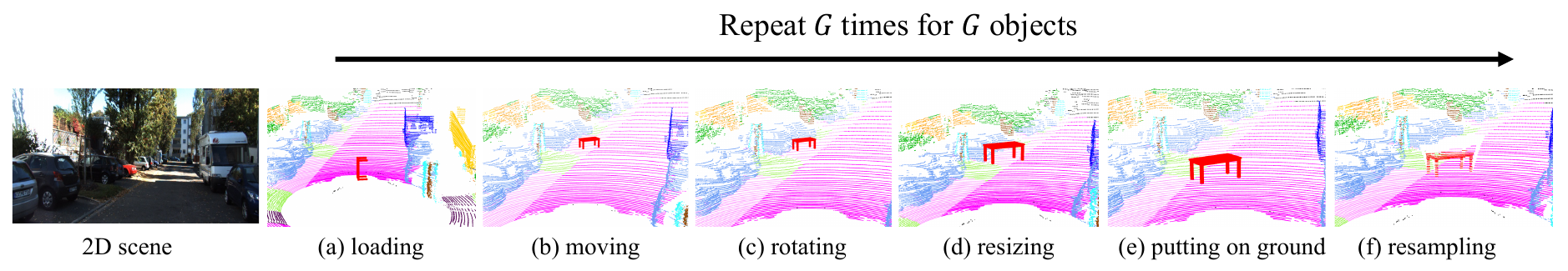}
\vspace{-1.2em}
\caption{Our outlier synthesis pipeline: (a) \textbf{loading} a ShapeNet object; (b) randomly \textbf{moving} it away from the scene center; (c) randomly \textbf{rotating} it around the gravity direction; (d) randomly \textbf{resizing} it; (e) \textbf{putting} it \textbf{on ground}; (f) \textbf{resampling} points on the object to blend into the scene. We repeat this pipeline on the fly $G$ times for inserting $G$ objects. }\label{fig:shapenet_synthesis}
\vspace{-1.2em}
\end{figure*}

\subsection{Point-wise Abstaining Penalty Learning}
\label{Point-wise abstaining mechanism}

\begin{figure*}[t]
\centerline{\includegraphics[width=0.7\textwidth]{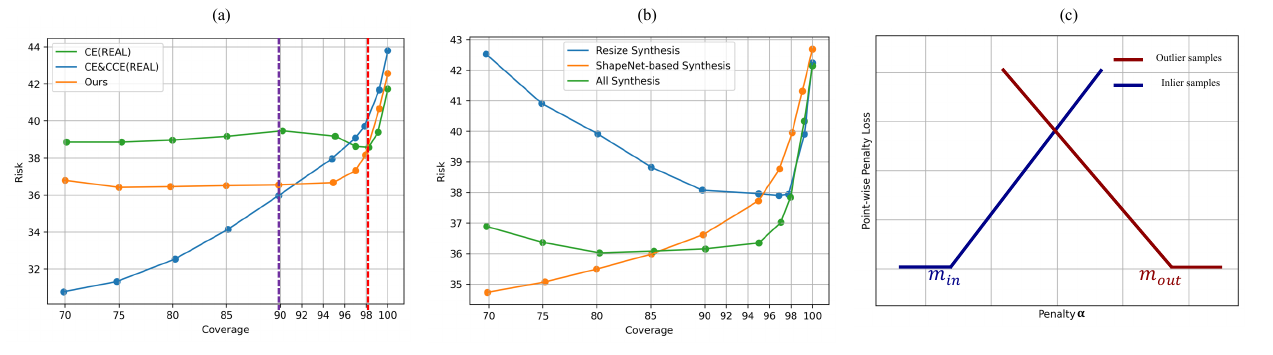}}
\vspace{-1em}
\caption{(a). Comparison with the SOTA using the Risk-Coverage curves; (b). Comparison with different outlier synthesis pipeline using the risk-coverage curve; (c). Relationship between point-wise penalty loss and penalty $\alpha$. }
\label{fig:sota_comparison_ablation_conceptua}
\vspace{-1.5em}
\end{figure*}

\textbf{Point-wise upgrade.} The reason why the disadvantages of CCE exceed its benefits is its sub-optimal calibration for the inlier probabilities $\mathbf{p^y}$ and the outlier probabilities $\mathbf{p^o}$. Inspired by `learning to abstain' \citep{liu2019deep}, where the calibration between the reject and non-reject options is adeptly handled, we propose our new learning paradigm. Specifically, \citet{liu2019deep} employs a fixed calibrating factor to achieve this. While this may suffice for image-level SC, it is inadequate for point-wise outlier detection. This is because, in LiDAR outlier detection, we need a different calibrating factor for each point to capture the subtle differences between them, such as between remote (sparse) and near (dense) points, as well as between inlier and outlier points. Therefore, we upgrade this fixed calibrating factor to a point-wise one, defined as abstaining penalty $\alpha$, and introduce a point-wise penalty loss to supervise the network to learn the subtle differences between various points, which can be expressed as follows:
\begin{equation}
\begin{gathered}
\mathbf{\alpha} := \left\{\alpha_i = - \log(\sum_{j=1}^{c} e^{\hat{y}_{i,j}}) | i = 1,\ldots, n\right\}\\
\ell^{\text{penalty}} := \frac{1}{m}\sum_{S_m}\frac{1}{n}\sum_{i =1}^n \Biggl\{\mathbb{I}(y_i \neq c+1)\text{max}(\alpha_i - m_{\text{in}},0) \\
+ \mathbb{I}(y_i = c+1)\text{max}(m_{\text{out}} - \alpha_i,0)\Biggr\}
  \end{gathered}
  \label{eq:penalty-formulation}
\end{equation}
As shown in \cref{fig:sota_comparison_ablation_conceptua}-(c), the hyperparameters $m_{\text{in}}$ and $m_{\text{out}}$ ensure that the inliers are associated with penalties lower than $m_{\text{in}}$, while the outliers exhibit penalties higher than $m_{\text{out}}$. Note that in our experiments, the penalties are negative and we set the value of $m_{\text{in}}$ to -12, and $m_{\text{out}}$ to -6.

Moreover, our new `learning to abstain' formulation for this task with this point-wise penalty is defined as follows:
\begin{equation}
\begin{gathered}
\ell^\text{abstain} := \\
\frac{1}{m}\sum_{S_m}\frac{1}{n}
     \sum_{i =1 }^n \Biggl\{ \underbrace{-\mathbb{I}(y_i \neq c+1)\log{ \left\{p^y_{i,y_i} + \underbrace{\frac{p^o_i}{ (- \alpha_i)^2 }}_{\text{abstaining term} } \right\}}}_\text{for inlier samples} \\
     \underbrace{-\mathbb{I}(y_i = c+1)\sum_{j=1}^{c} \log{ \left\{ p^y_{i,j} + \underbrace{\frac{p^o_i}{ (- \alpha_i)^2  }}_{\text{abstaining  term}}  \right\}}}_\text{for outlier samples} \Biggr\}
  \end{gathered}
  \label{eq:abstain-formulaiton}
\end{equation}
\textbf{Intuition.} 
Minimizing the point-wise penalty loss \cref{eq:penalty-formulation} results in assigning lower $\alpha_i$ to inliers, thereby leading to higher values of $(-\alpha_i)^2$. This effectively suppresses the contribution of $p^o_i$ and allows $p^y_{i,y_i}$ to play a dominant role in the point-wise abstain loss \cref{eq:abstain-formulaiton}. Likewise, higher $\alpha_i$ are allocated to outliers, resulting in lower values of $(-\alpha_i)^2$. This allows $p^o_i$ to play a dominant role and, consequently, suppresses the contribution of $\{p^y_{i,j} | j = 1,\ldots,c\}$ in the point-wise abstain loss \cref{eq:abstain-formulaiton}. Since this learning paradigm is defined in a point-wise manner, it has the potential to capture the subtle difference between inliers and outliers despite LiDAR point clouds lack semantically-rich information.

The total loss can be expressed as follows:
\begin{equation}
\begin{gathered}
\ell^\text{total} := \lambda^\text{abstain} \ell^\text{abstain} + \lambda^\text{penalty} \ell^\text{penalty}
   \end{gathered}
  \vspace{-0.5em}
  \label{eq:total loss formulation for static penalty }
\end{equation}
\subsection{Outlier Synthesis Pipeline}
\label{shapenet_anomaly}
\begin{table*}[t]
  \centering
   \adjustbox{width=0.65\textwidth}{
    \begin{tabular}{c|ccc|ccc}
    \toprule
    \multirow{2}{*}{Method} & \multicolumn{3}{c|}{SemanticKITTI \citep{behley2019semantickitti}}& \multicolumn{3}{c}{NuScenes \citep{caesar2020nuscenes} } \\
          & AUPR   & AUROC   & $\text{mIoU}_{\text{old}}$ & AUPR   & AUROC   & 
          $\text{mIoU}_{\text{old}}$ \\
    \midrule
    Closed-set C3D & - & - & 58.00 & - & - & 58.70 \\
    \midrule
    C3D + MSP \citep{hendrycksBaselineDetectingMisclassified2018} & 6.70 & 74.00 & 58.00 & 4.30 & 76.70 & 58.70 \\
    C3D + MaxLogit \citep{hendrycks2019scaling} & 7.60 & 70.50 & 58.00 & 8.30 & 79.40 & 58.70 \\
    C3D + MC-Dropout \citep{gal2016dropout} & 7.40 & 74.70 & 58.00 & 14.90 & 82.60 & 58.70 \\
    C3D + REAL \citep{cen2022open} & 20.08 & 84.90 & 57.80 \textcolor{red}{(0.20 $\downarrow$)} & 21.20 & 84.50 & 56.80 \\
    C3D + APF \citep{li2023open} & 36.10 & 85.60 & 57.30 \textcolor{red}{(0.70 $\downarrow$)} & - & - & - \\
    C3D + LiON (ours)& 
    44.68 \textcolor{red}{(8.58 $\uparrow$)} & 
    92.69 \textcolor{red}{(7.09 $\uparrow$)} & 
    57.56 \textcolor{red}{(0.44 $\downarrow$)} & 
    31.58 \textcolor{red}{(10.38 $\uparrow$)} & 
    95.24 \textcolor{red}{(10.74 $\uparrow$)} & 
    59.11 \textcolor{red}{(0.41 $\uparrow$)} \\
    \bottomrule
    \end{tabular}
    }
    \caption{Comparisons with previous methods. C3D refers to the base segmentation model, Cylinder3D \citep{zhu2021cylindrical}. Comparisons with APF on NuScenes are not conducted, as the results are unavailable and its code is not publicly released.}
  \label{tbl:comparison_to_sota}
\end{table*}
REAL synthesizes outliers by resizing the objects presenting in existing scene, as depicted in the left panel of \cref{fig:teaser}-(b). However, we observe that this synthesis pipeline fails to represent the real outlier in two aspects: 1) the limited variety of objects in the existing scene makes it challenging to compensate for the lack of semantically-rich information through resizing; 2) learning from these synthesized outliers may lead to a model that classifies real outliers solely based on point sparsity. As such, we resort to an additional dataset, ShapeNet, which consists of 220,000 models classified into 3,135 categories. As shown in \cref{fig:shapenet_synthesis}, we repeat the synthesis pipeline for $G$ times to insert $G$ outlier objects. For each object, there are six steps as follows:

(a) To synthesize diverse outliers, we first randomly decide the number of outlier objects $G$ according to a Binomial distribution\footnote{The Binomial(a, b) generates a random value from a binomial distribution with `a' trials and `b' probability of success per trial.}. The specific distribution used is  $\text{Binomial}(20,0.3)$. We then \textbf{load} $G$ objects from ShapeNet into the given scene $\mathbf{x}$, where the probability of not adding any object into $\mathbf{x}$ is also considered. 

(b) Then, for each object $\mathbf{s} \in \mathbb{R}^{l \times 3}$, we \textbf{move} $\mathbf{s} \in \mathbb{R}^{l \times 3}$ away from scene center $x^c$, alone the x-axis by $d^x \sim \text{Uniform}(r^\text{min}, 0.8 * r^{\rm max})$\footnote{The Uniform(a, b) generates a random value from a uniform distribution with lower bound `a' and upper bound `b'.} that $r^\text{min}$ is the distance of the closest point from $x^c$ and $r^\text{max}$ is the furthest point from $x^c$.

(c) Next, we \textbf{rotate} $\mathbf{s}$ around $x^c$ on the xy-plane (around the gravity direction) for $d^\text{lon} \sim \text{Uniform}(0, 360)$ degrees and denote the resulting object as $\mathbf{s} = (\mathbf{u}, \mathbf{v}, \mathbf{w})$. There is a probability that $\mathbf{s}$ does not overlap with $\mathbf{x}$, after moving and rotating. Therefore, if $\mathbf{s}$ is positioned outside of $\mathbf{x}$, the subsequent steps are not carried out and we move on the next object. Specifically, we stop synthesis process if:
\begin{equation}
\begin{gathered}
\min \left\{ |\bar{u} - i| + | \bar{v} - j | \right\} > \Delta, (i, j, k) \in \mathbf{x}
  \end{gathered}
  \label{eq:justification for overlap}
\end{equation}
Here, $\bar{u}$ and $\bar{v}$ represent the mean u and mean v of $s$, respectively. We set $\Delta$ to 1 in our experiments.

(d) Then, since the objects from ShapeNet tend to be smaller in size compared to those in the existing scene, we proceed to \textbf{resize} $\mathbf{s}$ by a factor of $k \sim \text{Uniform}(1,7)$. 

(e) Following this, we \textbf{put} $\mathbf{s}$ \textbf{on ground} by setting its last axis to $\mathbf{\Tilde{w}} = \mathbf{w} - \Delta_w$, where $\Delta_w$ represents the distance between the bottom of $\mathbf{s}$ and the point on $\mathbf{x}$ that are the closest to $\mathbf{s}$ along the gravity direction and falls into the x-y plane projection of $s$. The resulting object is $\mathbf{s} = (\mathbf{u}, \mathbf{v}, \mathbf{\Tilde{w}})$.


(f) Finally, to consider the realistic LiDAR's sampling pattern, we merge $\mathbf{s}$ into $\mathbf{x}$ by adjusting the radii of $\mathbf{x}$. Specifically, we represent $\mathbf{s}$ and $\mathbf{x}$ using spherical coordinates:
\begin{equation}
\begin{gathered}
\mathbf{s} := \{s_j = (\text{lon}_j, \text{lat}_j, \text{r}_j) | j = 1, \ldots,l\} \\
\mathbf{x} := \{x_k=(\text{lon}_k, \text{lat}_k, \text{r}_k) | k = 1, \ldots, n \}
  \end{gathered}
  \label{eq:spherical coordinates}
\end{equation}
Here, the lon represents longitude, lat represents latitude, and r represents radius. For each $x_k$, we replace $\text{r}_k$ with $\text{r}_j$ if $s_j$ satisfy $|\text{lon}_k - \text{lon}_j| < \Delta_\text{lon}$ and $|\text{lat}_k - \text{lat}_j| < \Delta_\text{lat}$. During our experiment, we set $\Delta_\text{lon}$ to $0.02$ and $\Delta_\text{lat}$ to $0.2$. When multiple $s_j$ satisfy the above criterion, we use their smallest r to replace $\text{r}_k$.

\subsection{Dynamic Penalty}
As shown in \cref{fig:teaser}-(b), the outliers synthesized through resizing are further from the inlier data distribution in terms of point sparsity compared to those synthesized by our pipeline. Hence, to maximize the benefits of the point-wise learning paradigm, we introduce a dynamic penalty loss that handles points in a customized manner:
\begin{equation}
\begin{gathered}
\ell^{\text{dynamic penalty}} := \\
\frac{1}{m}\sum_{S_m}\frac{1}{n} \sum_{i =1}^n  \Biggl\{\mathbb{I}(y_i \neq c+1 \text{ \& } y_i \neq c+2)) \text{max}(\alpha_i - \beta_{\text{in}} m_{\text{in}},0) \\
+ \mathbb{I}(y_i = c+1)\text{max}(\beta_{\text{rout}} m_{\text{rout}} - \alpha_i,0) \\
+ \mathbb{I}(y_i = c+2) \text{max}(  \beta_{\text{sout}} m_{\text{sout}} - \alpha_i,0)\Biggr\}
  \end{gathered}
  \label{eq:dynamic penalty loss}
\end{equation}
The $\{c+2\}$ denotes the outlier class label generated by ShapeNet. The weight parameter $\beta$ associated with the threshold $m$ is initialized as 1 and is learnable. In our experimental setting, we set the value of $m_{\text{sout}}$ to -7, and $m_{\text{rout}}$ to -6. Consequently, the total loss becomes:
\begin{equation}
\begin{gathered}
\ell^\text{total} := \lambda^\text{abstain} \ell^\text{abstain} + \lambda^\text{dynamic penalty} \ell^\text{dynamic penalty}
  \end{gathered}
  \vspace{-0.5em}
  \label{eq:total-loss-formulation-for-dynamic-penalty}
\end{equation}
\section{Experiments}
\label{sec:experiments}
\begin{figure*}[t]
\centerline{\includegraphics[width=0.95\textwidth]{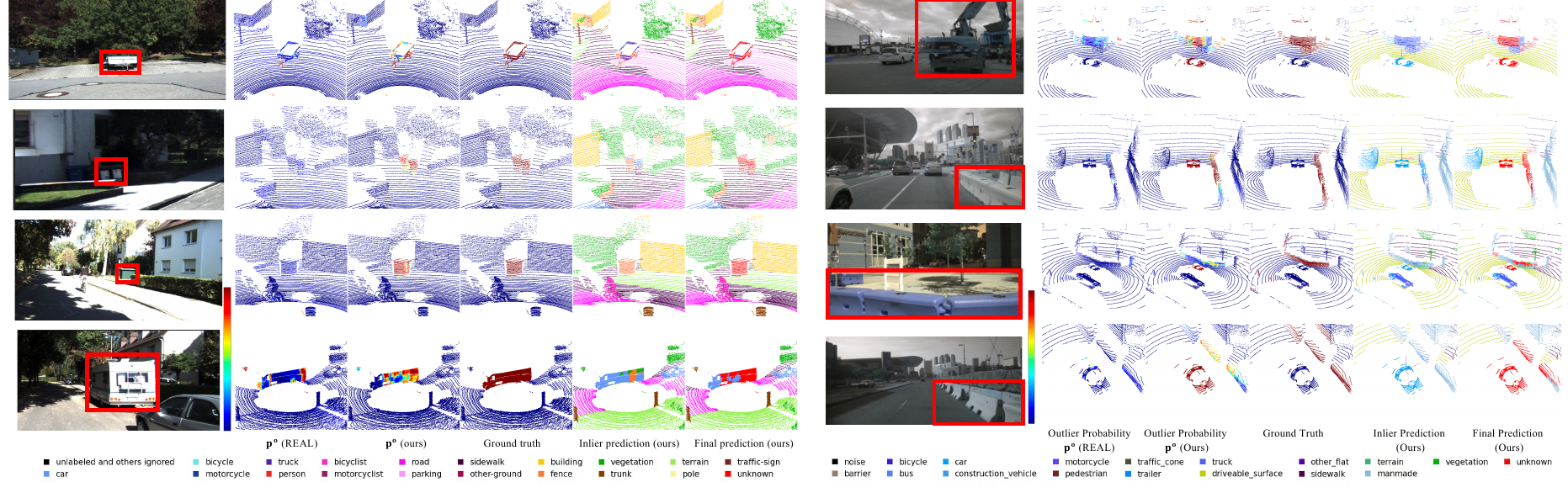}}
\vspace{-1em}
\caption{ Qualitative comparison results for SemanticKITTI (left) and NuScenes (right); Inlier prediciton indicates semantic segmentation; The final prediction is obtained by integrating the inlier prediction with $\mathbf{p^o}$ using a threshold of $0.5$.}
\label{fig:quatitative_merged}
\vspace{-1.em}
\end{figure*}

\subsection{Dataset}
\textbf{SemanticKITTI} \citep{behley2019semantickitti} is a driving-scene dataset designed for point cloud segmentation. The point clouds are collected using the Velodyne-HDLE64 LiDAR in Germany. The dataset consists of 22 sequences, with sequences 00 to 10 utilized as the training set, sequence 08 serves as the validation set, and sequences 11 to 21 used as the test set. After merging classes with different moving statuses and ignoring classes with a small number of points, 19 classes remain for training and evaluation. Consistent with prior work, we designate $\{\text{other-vehicle}\}$ as outlier class.

\textbf{NuScenes} \citep{caesar2020nuscenes} consists of 1000 scenes, each lasting 20 seconds, captured using a 32-beam LiDAR sensor, which leads to its challenging nature (a sparser LiDAR point cloud makes the classification task more difficult). It contains 40,000 frames sampled at 20Hz and has official training and validation splits. After merging similar classes and removing rare/useless classes including `ego-car', there are 16 remaining classes for training and evaluation. The classes designated as outliers include $\{\text{barrier},\text{constructive-vehicle},\text{traffic-cone}, \text{trailer}\}$.


\subsection{Evaluation Metric}
\label{section: evaluation metrics}

\textbf{Traditional evaluation metrics: } Consistent with previous work \citep{cen2022open}, we employ inlier mean intersection over union (mIoU$_\text{old}$) metric to evaluate the performance of inlier classification, while the AUPR and AUROC are utilized to assess the performance of outlier classification.

\textbf{New evaluation metrics: } The loss-based selective risk (\cref{eq:empirical-risk-formulation-definition}) is sub-optimal for serving as an evaluation metric for point-wise classification task because it is not sensitive to the class imbalance which is crucial in this task. Thus, we upgrade it into mIoU-based selective risk, as shown below:
\begin{equation}
\begin{aligned}
    \hat r(f,g|S_m) := \frac{ 100 - \text{mIoU}^{S_m|g }_\text{old}  }{\phi(g|S_m)}
  \end{aligned}
  \label{eq:empirical risk formulation1}
\end{equation}
Here, $\text{mIoU}^{S_m|g}_\text{old}$ represents the mIoU$_\text{old}$ calculated for the sub-dataset $S_m$ under the selective model condition $g$. Through this upgradation, the selective risk becomes more comparable, as $\hat{r} \in [0,100]$, and the mIoU inherently accounts for sensitivity to class imbalance. Furthermore, the definitions of selective AUPR and AUROC can be found in the appendix. With these new evaluation metrics, we can draw risk/AUPR/AUROC-coverage curves to obtain deeper understanding of model performance.

\subsection{Comparisons with State-of-the-art Methods}
We use Cylinder3D \citep{zhu2021cylindrical} as the baseline segmentation model to ensure a fair comparison. The computational cost remains low with the addition of an outlier detection head, achieving 7 fps on a single NVIDIA 3090 GPU.

\textbf{Quantitative comparison:} As illustrated in \cref{tbl:comparison_to_sota}, our method achieves a new SOTA in outlier class segmentation for SemanticKITTI and NuScenes, surpassing the previous SOTA by a significant margin. Specifically, our method achieves an AUPR of 44.68 and an AUROC of 92.69, which exceeds the previous SOTA scores by 8.58 and 7.09 in SemanticKITTI. Moreover, our method achieves an AUPR of 31.58 and an AUROC of 95.24 in NuScenes, which exceeds the previous SOTA scores by 10.38 and 10.74, respectively.

\textbf{Qualitative comparison:} Qualitative results, as illustrated in \cref{fig:quatitative_merged}, demonstrate that our method not only locates the outliers more accurately but also does so with greater confidence compared to REAL. Furthermore, in NuScenes (\cref{fig:quatitative_merged}-right ), our method accurately identifies the `ego-car' as outliers, although this category is not incorporated into the training and evaluation phases. These findings provide further evidence of the superiority of our method.



\textbf{Comparison between LiON and the previous SOTA (APF) from another perspective:} In addition to better quantitative results, LiON is easier to implement and requires less computational cost compared to APF. LiON can be implemented by simply adding an extra classifier to an arbitrary LiDAR-based segmentation network and training the network in a single stage using our novel learning paradigm and two different outlier synthesis pipelines. In contrast, APF requires not only an arbitrary segmentation network but also several learnable prototypes, a prototypical constraint module, a generator, a discriminator, and an adversarial mapper, which significantly increase computational costs. Further, APF relies on a two-stage training process, complicating the reproducibility of its results.

\textbf{Risk-Coverage curve comparison:} As shown in \cref{fig:sota_comparison_ablation_conceptua}-(a), the problem of the CE\&CCE, analyzed above, is reflected as a high risk at high coverages, while the CE exhibits an unstable trend when coverage decreases. In contrast, our method achieves a highly competitive risk compared to them at high coverages. With decreasing coverage, our method shows a consistent decline to a plateau in risk.

\textbf{Why our method achieve a higher risk compared to REAL when coverage is below 90\%? (purple dotted line)} This is because our method rejects most real outlier samples/points when coverage exceeds 90\%. This is proved by that the threshold set to achieve 90\% coverage is 0.0051, indicating that all samples with a predicted outlier probability higher than 0.0051 are rejected. This means that when coverage is around 90\%, the outlier probabilities of the remaining samples are relatively small, and there are few real outliers left. Therefore, to further decrease the coverage, our method tends to reject more true inliers than true outliers.


However, we believe that robust performance in high coverage is more critical than in low coverage due to the fact that the outlier objects are rare in the real world.


\subsection{Ablation Study}
\begin{table}[t]
  
  \centering
  \begin{adjustbox}{width=0.35\textwidth}
        \begin{tabular}{c|c|c|c|c}
        \hline
        Penalty & Dynamic Penalty &  AUPR $\uparrow$ & AUROC $\uparrow$ & $\text{mIoU}_{\text{old}} \uparrow$ \\
         \hline
         \hline
         \checkmark  & & 43.69 & 92.51 & 57.47 \\
          & \checkmark & \textbf{44.68} & \textbf{92.69} & \textbf{57.56} \\
        \hline
        \end{tabular}
    \end{adjustbox}
    \caption{ Ablations for dynamic penalty setting in SemanticKITTI.}
    \label{tab:Ablation for semantic kitti } 
\vspace{-0.5em}
\end{table}

\textbf{Effectiveness of dynamic abstaining penalty.} As demonstrated in \cref{tab:Ablation for semantic kitti }, the penalty setting is denoted by \cref{eq:total loss formulation for static penalty }, while the dynamic penalty setting is represented by \cref{eq:total-loss-formulation-for-dynamic-penalty}. The dynamic penalty setting achieves the best performance. These results provide evidence for the effectiveness of our dynamic penalty design.

\begin{table}[t]
  \centering

  \centering
  \begin{adjustbox}{width=0.35\textwidth}
    \begin{tabular}{c|c|c|c|c}
    \toprule
     ShapeNet & Resize & AUPR & AUROC & $\text{mIoU}_{\text{old}}$ \\
     \midrule
     \midrule
    \checkmark & & 29.14 \textcolor{red}{(14.55 $\downarrow$)} & 89.56 \textcolor{red}{(2.95 $\downarrow$)}& 57.31 \textcolor{red}{(0.16 $\downarrow$)} \\
     & \checkmark & 41.82 \textcolor{red}{(1.87 $\downarrow$)} & 93.04 \textcolor{red}{(0.53 $\uparrow$)} & 57.30 \textcolor{red}{(0.17 $\downarrow$)} \\
    \checkmark & \checkmark & \textbf{43.69} & \textbf{92.51} & \textbf{57.47} \\
    \bottomrule
    \end{tabular}
    \end{adjustbox}
      \caption{ Ablation study on outlier synthesis pipeline in SemanticKITTI.}
  \label{tab:Ablation for synthesis method}
  \vspace{-1em}
\end{table}

\textbf{Effectiveness of outlier synthesis pipeline.} As shown in \cref{tab:Ablation for synthesis method}, excluding the ShapeNet synthesis pipeline results in a decrease in AUPR by 1.87 and an increase in AUROC by 0.53. Moreover, the AUPR and AUROC drop significantly by 14.55 and 2.95, respectively, without the resize synthesis pipeline. This raises the question of \textbf{whether ShapeNet synthesis pipeline is trivial}.

There are two types of real outliers: those that are distant from the inlier distribution (referred to as far real outliers) and those that lie closer to the inlier distribution (referred to as near real outliers). As shown in \cref{fig:sota_comparison_ablation_conceptua}-(b), the resize synthesis pipeline shows an initial decrease in risk followed by an increase. This occurs because this pipeline can approximate the far real outlier distribution but struggles with approximating the near real outlier distribution. 

On the other hand, our ShapeNet synthesis pipeline exhibits a consistent reduction in risk as the coverage decreases, as it can synthesize outliers that approximate both near and far real outlier distributions.  The former result from considering the LiDAR sampling pattern, while the latter stem from considering the diversity of real outliers.

\section{Conclusion}
In this work, we first revisit previous methods using the unified lens of selective classification and propose a new formulation which effectively captures the subtle differences between inliers and outliers. Then, we design a novel outlier synthesis pipeline to synthesize diverse and realistic outliers, compensating for the lack of semantically-rich information in point clouds. Experimental results demonstrate the superiority of our method across both traditional outlier detection metrics and newly introduced metrics. Our method has achieved SOTA performance on public benchmark datasets.

\bibliography{aaai25}

\end{document}